\documentclass[conference]{IEEEtran}
\IEEEoverridecommandlockouts
\usepackage{cite}
\usepackage{amsmath,amssymb,amsfonts}
\usepackage{algorithm}
\usepackage{underscore}
\usepackage{algpseudocode}
\usepackage{graphicx}
\usepackage{textcomp}
\usepackage{xcolor}
\usepackage{geometry}
\usepackage{booktabs,siunitx}
\usepackage{capt-of}  
\usepackage{cuted}    
\usepackage{lipsum}
\usepackage[english]{babel}
\algnewcommand\algorithmicforeach{\textbf{for each:}}
\algnewcommand\ForEach{\item[ \algorithmicforeach]}

\def\BibTeX{{\rm B\kern-.05em{\sc i\kern-.025em b}\kern-.08em
    T\kern-.1667em\lower.7ex\hbox{E}\kern-.125emX}}

\begin{document}

\title{Landmark-based Localization using Stereo Vision and Deep Learning in GPS-Denied Battlefield Environment\\
}
\author{\IEEEauthorblockN{Ganesh Sapkota}
\IEEEauthorblockA{\textit{Department of Computer Science} \\
\textit{Missouri University of Science and Technology}\\
Rolla, MO, USA \\
gs37r@mst.edu}
\and
\IEEEauthorblockN{Sanjay Madria}
\IEEEauthorblockA{\textit{Department of Computer Science} \\
\textit{Missouri University of Science and Technology}\\
Rolla, MO, USA\\
madrias@mst.edu}
}
\maketitle 
\begin{abstract}
Localization in a battlefield environment is increasingly challenging as GPS connectivity is often denied or unreliable, and physical deployment of anchor nodes across wireless networks for localization can be difficult in hostile battlefield terrain. Existing range-free localization methods rely on radio-based anchors and their average hop distance which suffers from accuracy and stability in dynamic and sparse wireless network topology. Vision-based methods like SLAM and Visual Odometry use expensive sensor fusion techniques for map generation and pose estimation. This paper proposes a novel framework for localization in non-GPS battlefield environments using only the passive camera sensors and considering naturally existing or artificial landmarks as anchors. The proposed method utilizes a custom-calibrated stereo vision camera for distance estimation and the YOLOv8s model, which is trained and fine-tuned with our real-world dataset for landmark recognition. The depth images are generated using an efficient stereo-matching algorithm, and distances to landmarks are determined by extracting the landmark depth feature utilizing a bounding box predicted by the landmark recognition model. The position of the unknown node is then obtained using the efficient least square algorithm and then optimized using the L-BFGS-B (limited-memory quasi-Newton code for bound-constrained optimization) method.
Experimental results demonstrate that our proposed framework performs better than existing anchor-based DV-Hop algorithms and competes with the most efficient vision-based algorithms in terms of localization error (RMSE). 
\end{abstract}
\section{ Introduction}
Traditional localization methods such as GPS rely on satellite signals, which can be easily disrupted or blocked in environments with dense vegetation, urban canyons, intentional jamming, or disrupted by adversaries. In the modern battlefield, where the use of GPS is often denied or unreliable, accurate and robust localization systems are crucial for situational awareness and effective mission planning\cite{wsn_battlefield_application,battlefield_surv_using_WSN}. Localization in a GPS-denied battlefield environment is challenging due to several factors. GPS relies on satellite signals to determine a user's location. 
\begin{figure}[!ht]
\centering
\includegraphics[width=0.4\textwidth]{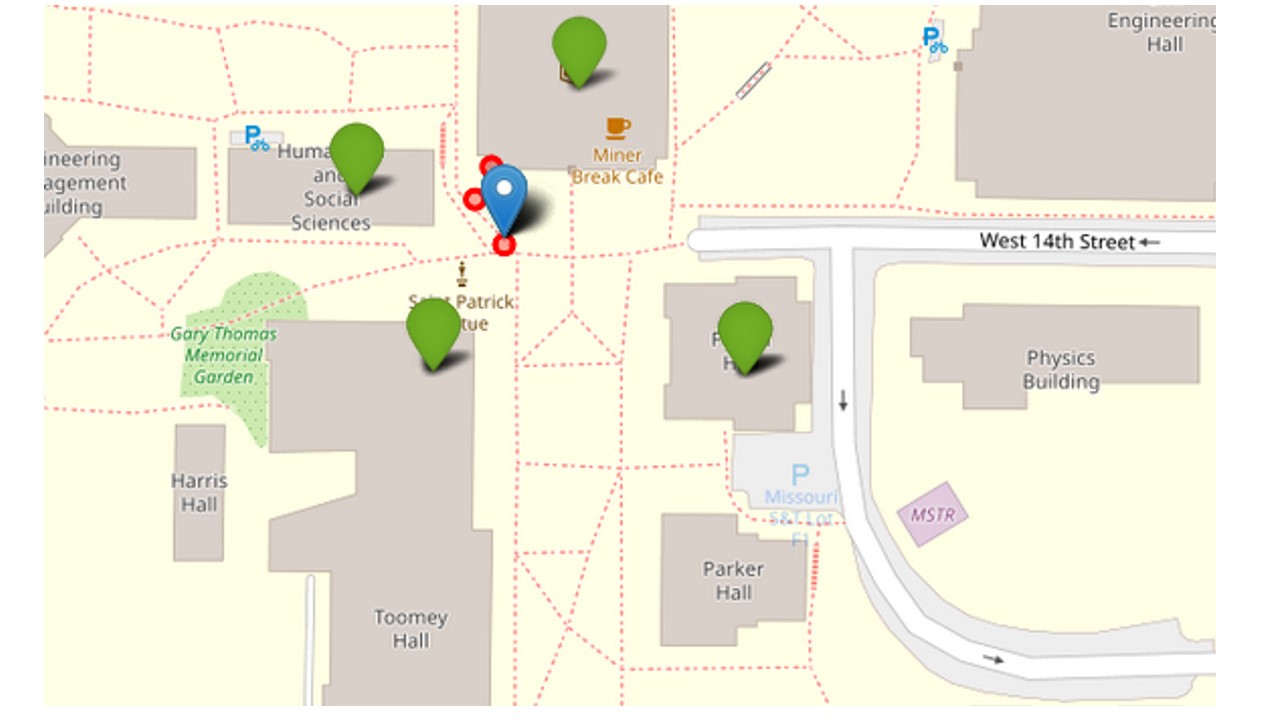}
\caption{Visualization of landmark anchor-based localization on a map. The green pin indicates the graphical landmarks detected by the landmark recognition model. The blue pin indicates the actual position of a mobile node and red solid circles indicate the estimated positions of the node using the proposed framework} 
\label{fig_localization}
\end{figure}
In battlefield environments, satellite signals can be intentionally jammed or disrupted by adversaries, making GPS unreliable or unusable. Battlefield environments are often characterized by electromagnetic interference and signal attenuation, making it difficult to establish reliable wireless communication links. This can make it challenging to deploy and maintain wireless sensor networks (WSNs), which are often used for localization and tracking\cite{battlefield_wsn} in GPS-denied environments. The physical deployment of anchor nodes, which are commonly used for localization in WSNs, can be difficult in hostile battlefield terrain. Anchor nodes often need to be placed in specific locations to provide accurate localization, but these locations may be inaccessible or dangerous\cite{anchor_node_localization}. So the deployments are often done through aerial scattering from airplanes, guided missiles, or balloons which makes the localization problem more complicated. Also, Battlefield environments are constantly changing, with moving troops, vehicles, and obstacles. This can make it difficult to maintain accurate localization information, as anchors and reference points may change or disappear. Adversaries may intentionally deploy false anchors or manipulate sensor data to confuse or mislead friendly forces. This can make it difficult to distinguish between real and fake anchors and can lead to inaccurate localization. Due to these challenges, localization in GPS-denied battlefield environments is an active area of research. Researchers are developing new methods and technologies to overcome these challenges and provide reliable localization for military operations.As an alternative localization scheme for GPS denied environment, we have proposed a landmark anchor-based localization framework using an efficient stereo vision method and a robust landmark recognition model using deep learning to estimate the precise position of an unknown mobile node without using radio-based communication as shown in Fig.\ref{fig_localization}.//
In our Landmark-based localization approach, we utilize landmarks, which are naturally existing or human-made objects with well-defined locations in battlefield regions, as reference points(anchors) to estimate the position of vehicles, forces, or a commander who guides the forces along the safe path. This approach has the advantage of being independent of active sensors that use excessive radio signals and can be applied in a wide range of environments. Passive sensors like the camera are less in weight and require low power to operate which makes it easier and feasible to mount on resource-constrained mobile devices\cite{passive_vision_sensor}.  We used the Stereo Vision\cite{stereo_vision1,stereo_vision3} technique that utilizes two cameras to capture a 2D image of a 3D scene and construct its depth map. Deep learning has revolutionized object detection and recognition, making it well-suited for landmark identification in complex environments. By combining stereo vision and deep learning, landmark-based localization can achieve high accuracy and robustness in GPS-denied environments. Stereo vision provides a rich source of information necessary to accurately estimate the distance to landmarks, while deep learning enables the identification and localization of landmarks with high precision. Combining the distance information from the stereo vision method and the landmark location information from the deep learning model we applied the trilateration algorithm to estimate the position of the node. The node might be the troop commander or the Vehicle carrying troops equipped with a stereo-vision camera.\\
Experimental results show that our landmark anchor-based localization framework outperforms the existing anchor-based DV-Hop methods and Vision-based pose estimation schemes.\\
This paper is organized as follows: Section II discusses Related Works about Vision-based localization schemes and range-free localization schemes such as DV-HOP methods. Section III describes our approach and explains the system overview in detail. Section IV discusses the Experimental setup, Dataset, performance results, and comparison with existing works. Section V concludes the paper with the future direction of research.

\section{Related Works}
\begin{figure}[!ht]
\centering
\includegraphics[width=0.4\textwidth]{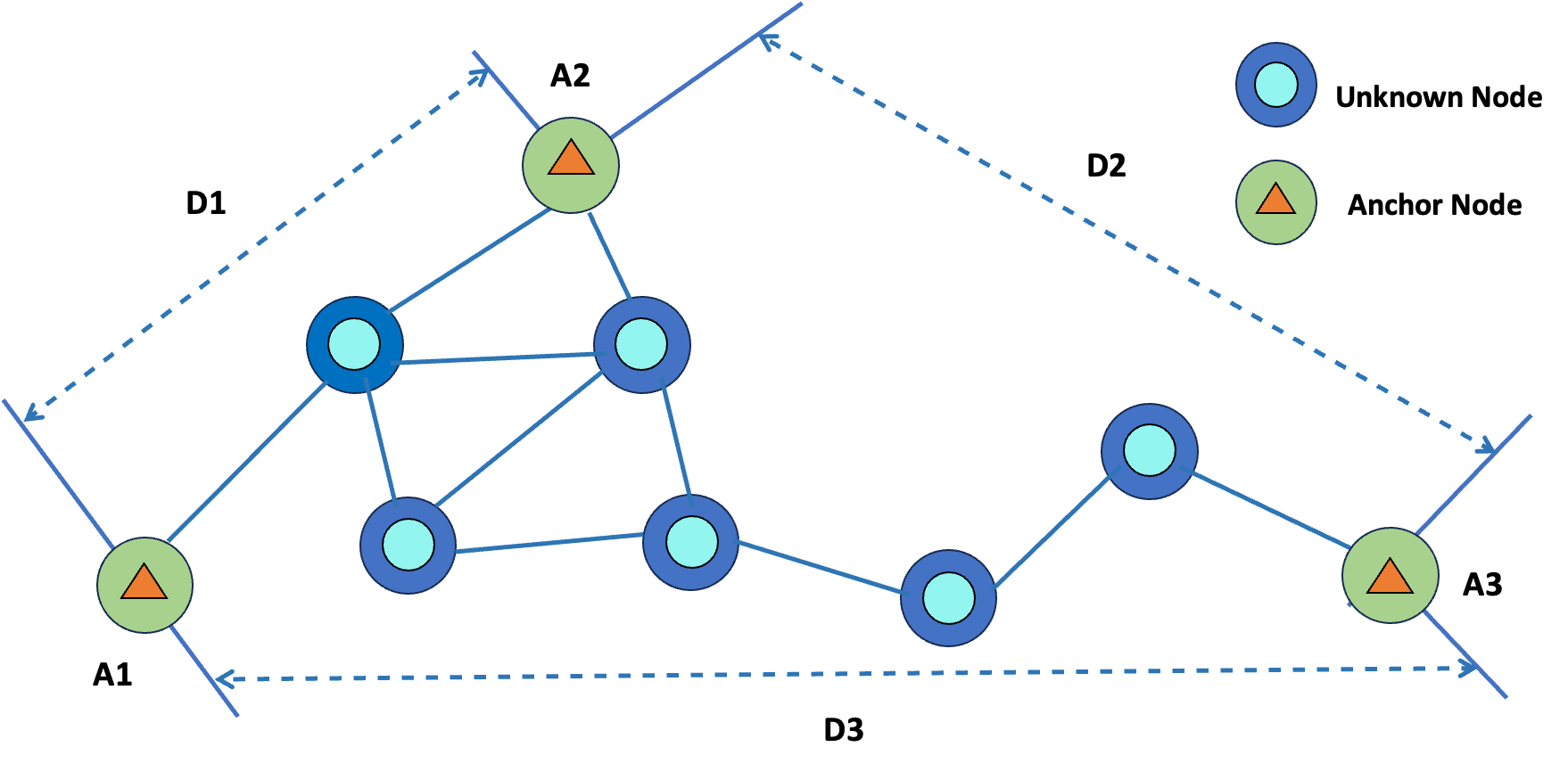}
\caption{Overview of DV-Hop localization algorithm} 
\label{fig_architecture_dvhop}
\end{figure}

Vision-based methods\cite{ref_slam,ref_orbslam,reforbslam2,orb_slam_object_detection}, for robot localization, have gained significant traction in recent years due to their ability to provide accurate and reliable positioning in environments where GPS signals are unavailable or unreliable. These methods utilize image or video data captured by cameras to determine the robot's location within its surroundings. Simultaneous Localization and Mapping (SLAM) technique ~\cite{ref_slam},  allows a mobile robot to simultaneously estimate its position and orientation (pose) within an environment while creating its map. However, it requires a fusion of cameras, lidar, or radar sensors to generate maps. The robot's state is represented by its pose while the map represents specific aspects of the environment such as the location of landmarks or obstacles.\\
ORB-SLAM~\cite{ref_orbslam}, utilizes FAST features for efficient image tracking and bundle adjustment\cite{ref_BA}, for map optimization.LSD-SLAM~\cite{ref_lsd_slam}, employs line segments as features for SLAM, enabling efficient localization in low-texture environments. These approaches are particularly effective in controlled indoor environments with distinct landmarks. SLAM techniques utilize a camera to extract features from images to track a robot's pose in relation to the map it constructs. However, if a moving object enters the field of view of the SLAM system, it might be misinterpreted as a stationary environmental feature causing inaccuracies in the map and the device's pose estimations. Over time, these errors may accumulate, potentially causing the robot to lose its trajectory, risking mission failure.\\
Visual Odometry (VO) algorithms\cite{deepLocalization,kittiVO}, estimate the robot's motion by tracking the changes in images or video frames captured as the robot moves. This method is suitable for continuous localization without the need for a pre-built map. Structure-from-Motion (SfM)\cite{ref_sfm}, the algorithm reconstructs the 3D structure of the scene from multiple images, allowing for accurate localization and pose estimation. This method is particularly useful for outdoor environments with well-defined features. Appearance-based Localization methods match captured images or video frames with a database of reference images to determine the robot's location. This approach is effective in environments with distinctive visual characteristics.\\
Deep Learning-based Localization\cite{ref_pose_estimate_yolo}, particularly convolutional neural networks (CNNs), have demonstrated impressive performance in vision-based localization. CNNs can extract high-level features from images and videos, enabling robust localization in various environments.\\
DV-hop-based range-free localization algorithms commonly used in wireless sensor networks (WSNs)\cite{dvhop_aps,dv_hop_Cyclotomic,dvhop_manhattanD,csdvhop,dvhop_nsgaii,dvhop_ss}, have received enormous attention due to their low-cost hardware requirements for nodes and simplicity in implementation. 
The algorithm estimates the location of unknown nodes based on their hop count to known anchor nodes. DV- works Algorithm starts with 3-stage\cite{dvhop_aps,dvhop_manhattanD} as shown in Fig.\ref{fig_architecture_dvhop}
Firstly, anchor nodes with known coordinates broadcast their positions and hop count (initially set to 1) to their neighboring nodes.
Secondly, unknown nodes receive the broadcasts from anchor nodes and update their hop count (hops). The hop count serves as a representation of the number of hops from an unknown node to the anchor node.
At stage three, unknown nodes estimate their positions based on the shortest distance to anchors and the location information of anchor nodes using the trilateration methods discussed in\cite{least_square1,least_square2,least_square3}
\begin{equation}
\label{averageHop}
hop_{avg} =  \frac{\sum_{{i}\neq{j}}^{n}\sqrt{(x_i - x_j)^2 + (y_i-y_j^2)}}{\sum_{{i}\neq{j}}^{n}hopCount_{i,j}}
\end{equation}
\begin{equation}
\label{distanceToanchor}
distance(d_{i,n}) = hop_{avg} * hopCount_{i,n}
\end{equation}
However, they have several drawbacks. The most prominent disadvantage is that node to anchor distance$(d_{i,n})$ in Eq.\ref{distanceToanchor} depends on the average distance per hop $(hop_{avg})$ as in Eq.\ref{averageHop}  and hop count estimation $hopCount_{i,n}$ instead of straight-line distances, which contributes to higher ranging error\cite{dv_hop_Cyclotomic}. High errors in anchor distance calculation directly impact in localization accuracy of unknown nodes. Also, the DV hop algorithm suffers from localization accuracy and stability when the network is dynamic and sparse\cite{dvhop_ss}. Sparse network topology contains multiple disconnected local networks that cannot exchange data and update the DV hop table effectively. Dynamic WSNs that contain mobile nodes, require an anchor to broadcast periodically which requires excessive energy and bandwidth requirements.
Since Battlefield WSNs can be highly unstable, dynamic, and sparse, the DV Hop-based method might be costly as well as ineffective.
\section{Our Approach}
Anchor node deployment in WSNs has been a challenging issue, especially in harsh environments like battlefields where network topology is sparse and dynamically changing. The sparse network contains multiple disconnected local network communities that can't exchange data with each other and update the DV-hop table. In a dynamic wireless network that contains mobile nodes, accurate DV-Hop updating requires the anchors to broadcast the beacon packets periodically and the network nodes need to forward the beacon packets which costs excessive energy and bandwidth consumption. So, many real-world applications fail to maintain physical anchors and set up the DV-Hop table for all devices through anchors' broadcast.
To avoid the difficulties of deploying and maintaining physical anchors in battlefield WSNs, our research proposes a new approach that utilizes existing physical and geographical landmarks as anchors which are first detected and then localized using a deep learning model and used to trilaterate the position of unknown sensor node. By employing landmarks anchors, the reliability and accuracy of localization are enhanced, as these physical landmarks are more resilient and durable than radio-based anchor nodes. Utilizing landmarks as anchors and employing the distance between sensors and landmarks as virtual coordinates expand the applicability of the proposed framework to sparse sensor networks deployed across extensive geospatial areas.
\begin{figure}[!ht]
\includegraphics[width=0.5\textwidth]{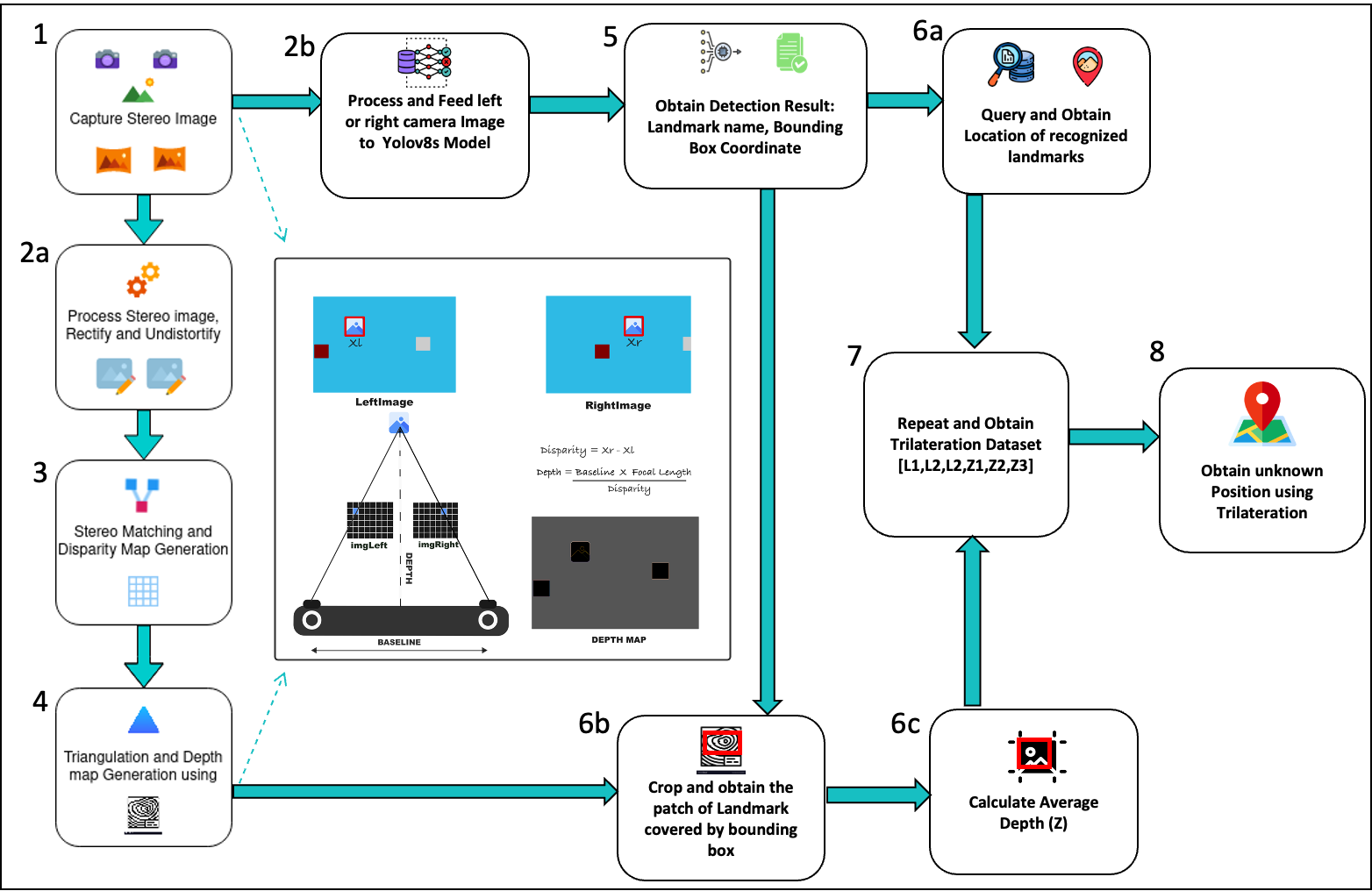}
\centering
\caption{System Overview: Distance estimation and landmark recognition steps and the results after two operations are fused to calculate the position of an unknown node.} 
\label{fig_architecture}
\end{figure}
Fig.\ref{fig_architecture} shows the system overview of our proposed Stereo Vision and deep learning-based localization framework which is summarized in the following steps:
\begin{enumerate}
    \item  A moving Node carrying a stereo rig captures stereo images of a scene containing a landmark Anchor
    \item 
    \begin{enumerate}
    \item Captured stereo images are processed using image processing techniques (Normalization, White balancing, Sharpening) and they are rectified and undistorted using camera calibration parameters.
    \item Processed images in step 2(a) from either the left or right camera are used as input to our landmark detection Model(YOLOv8s).
    \end{enumerate}
    \item Rectified images from step 2(a) are fed to a stereo-matching algorithm and a disparity map is generated for a given stereo pair.
    \item From disparity map, the depth map is generated using Eq.\ref{triangulation}
    \item YOLOv8s based Landmark recognition model gives the result containing landmark label and bounding box coordinates of the recognized landmark.
    \item 
     \begin{enumerate}
     \item  Based on the landmark label obtained in step 5, a pre-defined landmark database stored in a local device is queried to obtain the location.
     \item Bounding box coordinates obtained in step 5 are used to crop the landmark region patch from the depth map obtained in step 4.
     \item Depth levels of the new depth map obtained in step 6(b) are aggregated to obtain average distance $d_i$ to the detected landmark $l_i$.
     \end{enumerate}
     \item Process from Steps 1 to 6 are repeated to obtain trilateration tuple $({loc}_i, di)$ for each landmark anchor $l_i $ contributing to trilateration. Scanning the Scene is not terminated until the trilateration data $[l_1,l_2,l_3,d_1,d_2,d_3]$ are obtained.
     \item Using trilateration $[l_1,l_2,l_3,d_1,d_2,d_3],$ localization is done using the least square method and optimized the obtained position using the $L-BFGS-B$ optimization method. 
\end{enumerate}

\subsection{Stereo Vision Method}
We applied the stereo vision method to calculate the distance between an unknown node and the landmark anchor. Unlike the DV Hop method which uses the hop count and the average one-hop distance to the anchor node, we directly calculate the distance to the landmark anchors using stereo vision techniques.
\begin{figure}[!ht]
\centering
\includegraphics[width=0.25\textwidth]{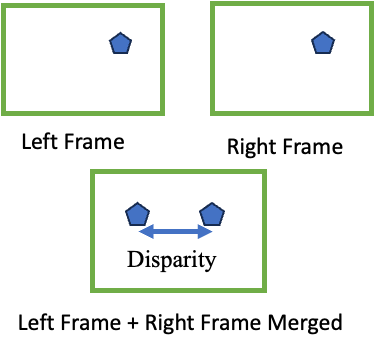}
\caption{Demonstration of Disparity in Stereo image Pair} 
\label{fig_disparity}
\end{figure}
\begin{figure}[!ht]
\centering
\includegraphics[width=0.45\textwidth]{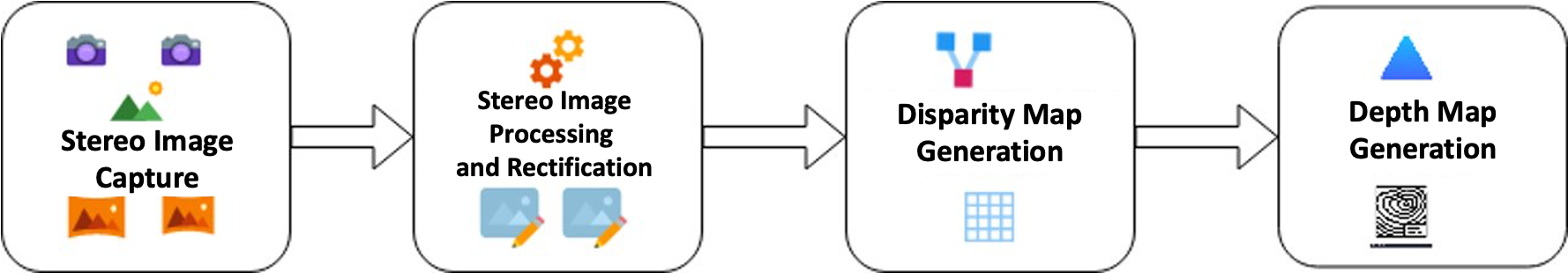}
\caption{Disparity and Depth Map Generation using Stereo Matching} 
\label{fig_depth_map}
\end{figure}
Stereo vision is a method of using two cameras to estimate the three-dimensional structure of a scene. It works by comparing the images of the same scene captured from two identical cameras separated by a horizontal distance (baseline) to find corresponding points. The point correspondence problems in our stereo images a solved using the Semi-global Block Matching (SGBM) \cite{sgm_original} that applies the sum of Squared Differences (SSD) as a cost function to ensure the best match.
We implemented the SGBM \textbf{Algorithm} \ref{alg_SGBM} and \textbf{Algorithm} \ref{alg_SSD} to generate a disparity map and apply triangulation to eventually obtain the depth map as shown in Fig.\ref{fig_depth_map}. Consider a stereo pair(leftImg,rightImg) captured from the left and right camera. Due to the horizontal separation of the camera by a baseline(B), an image point u(x,y) in the left frame gets shifted by a distance in the right frame and the point is found at(x,y) in the right frame. This shifted distance is referred to as the disparity(D) and as demonstrated in  Fig. \ref{fig_disparity}.
\begin{equation}
\label{calc_disp}
D = X_{leftImg}- X_{rightImg}
\end{equation}
\begin{equation}
\label{triangulation}
Z = f_{pixel} * B / D
\end{equation}
A disparity map is generated using Eq. \ref{calc_disp} and the intensity value at each pixel position represents the disparity of that pixel element as shown in Fig.
Now, the depth(Z) of each pixel element is calculated based on pixel intensities in the disparity map to generate a depth map using Eq.\ref{triangulation}, where f is the focal length of the camera.
Eq.2 indicates that disparity is inversely proportional to depth. This means that the disparity between corresponding image points increases when the scene point approaches nearer to the camera. Disparity calculation is a crucial step in stereo vision algorithms, as it provides the necessary information to estimate the depth of objects in the scene.
Then, we utilize the bounding box coordinates obtained from the landmark detection model to extract the depth patch and the depth features of the landmark. The depth patch comprises, the depth features of the detected landmark, which is further aggregated to calculate the average depth to the landmark anchor. This step is particularly important in obtaining a more precise distance calculation to the landmark.

\begin{algorithm}[H]
\caption{Semi-Global Block Matching (SGBM) Algorithm}
\label{alg_SGBM}
 \begin{algorithmic}[1]
\Require Stereo images $I_l$ and $I_r$
\Require Block size $blockSize$
\Require Disparity range $minD \ldots maxD$
\Require Discount factor $\alpha$
\Ensure Disparity map $D$

\State Initialize disparity map $D$ to zero
\State Initialize cost matrices $C_{l}$ and $C_{r}$ to zero
\State \textbf{for each} disparity value $d$ in $minD \ldots maxD$
\State Compute the sum of squared differences (SSD) between the block in $I_l$ and the corresponding block in $I_r$ with disparity $d$
\State Compute the cost function $C_d$ based on the SSD and discount factor $\alpha$
\State Update the cost matrices $C_{l}$ and $C_{r}$ based on the cost function $C_d$
\State Find the disparity value $d$ that minimizes the cost function $C_d$ at each pixel in the block
\State Update the disparity map $D$ by selecting the disparity value that minimizes the cost function $C_d$
\textbf{end for}
\end{algorithmic}
\end{algorithm}

\begin{algorithm}[H]
\caption{Sum of Squared Differences (SSD)}
\label{alg_SSD}
\begin{algorithmic}[1]
\Require Reference image $I_r$
\Require Template image $I_t$
\Require Displacement vector $d$
\Ensure Matching score $S$
\State Initialize matching score $S$ to zero
\State \textbf{for each} pixel $p$ in ${I_t}$
\State Compute the intensity difference between pixel $p$ in $I_t$ and the corresponding pixel $p+d$ in ${I_r}$
\State Square the intensity difference and add it to the matching score $S$

\State Normalize the matching score $S$ by dividing it by the number of pixels in $I_t$
\textbf{end for}
\end{algorithmic}
\end{algorithm}

\subsection{Landmark Detection using CNN}
For any localization algorithm, obtaining the position of anchors is a crucial step. In our approach, we used Deep Convolutional Neural Networks (CNN) to identify and obtain the location information of landmark anchors. Since we rely on only the camera sensors and captured stereo images for the localization of the node, we chose to apply recognition and localization of the landmark anchor using the YOLOv8-based object detection model\cite{yolov8_github} that divides an image into a grid and predicts bounding boxes and class probabilities for each grid cell. The ability to detect objects in a single pass and at different scales makes YOLO well-suited for identifying larger objects efficiently with fewer localization errors.
\begin{figure}[!ht]
\includegraphics[width=0.45\textwidth]{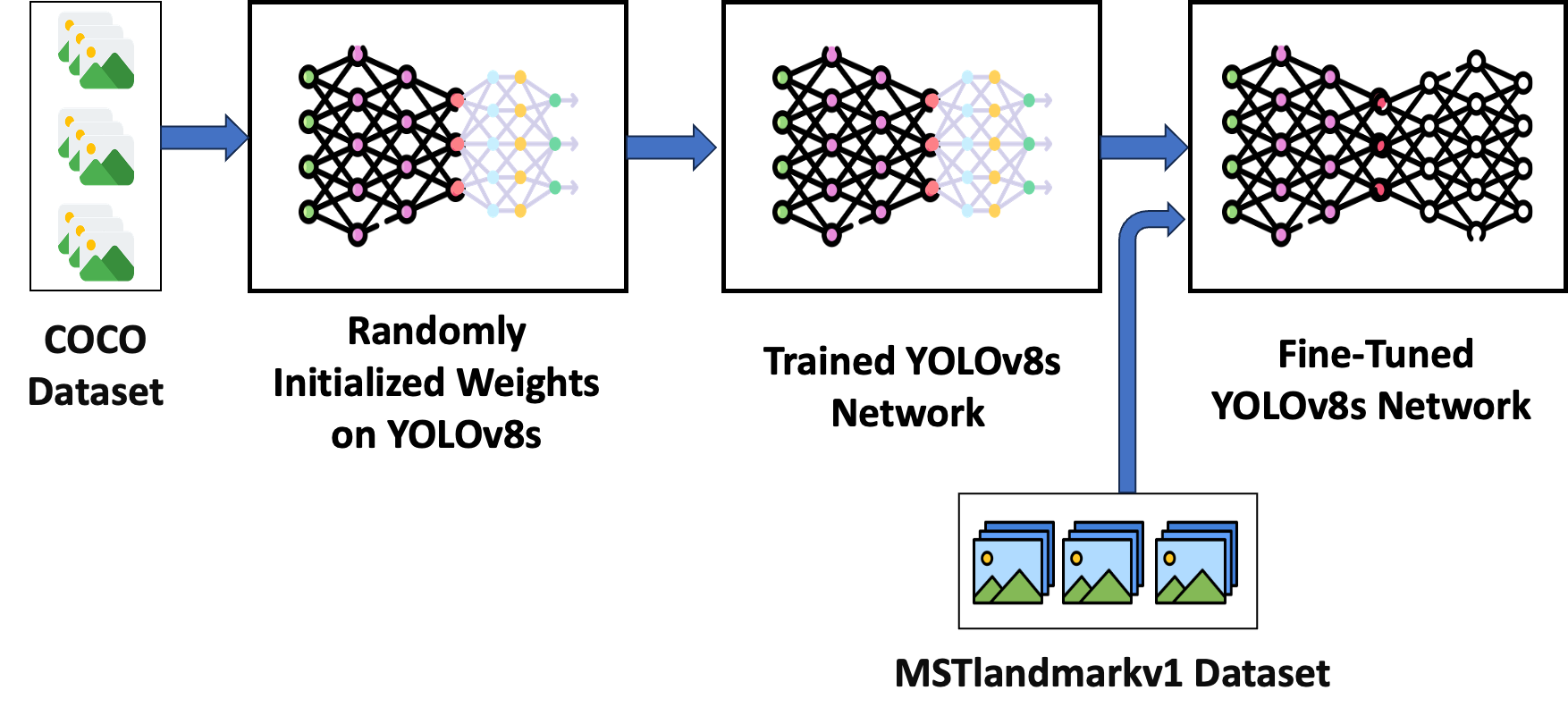}
\centering
\caption{Fine Tuning YOLOv8s with Custom Dataset} 
\label{fig_yolo_network}
\end{figure}
YOLOv8 uses the YOLOv8CSPDarknet backbone and C2f module (cross-stage partial bottleneck with two convolutions) to combine high-level features with contextual information to improve detection accuracy\cite{yolo_review}. In the output layer, sigmoid functions are used as the activation function for the objectness score, representing the probability that the bounding box contains an object. SoftMax functions are used for the class probabilities, representing the objects’ probabilities belonging to each possible class. CIoU and DFL functions are used for bounding box loss while cross-entropy for classification loss. We selected pre-trained YOLOv8s and re-trained the model with our landmark dataset to fit our recognition task. The model detects and recognizes the landmark in real-time with an average speed of $0.7ms$ for pre-process, $13.1ms$ inference, and $1.7ms$ post-process per image. The detection result gives the instance of the recognized landmark, which is used to obtain its geographic location by querying the predefined landmarks database and the bounding box coordinates, which are used to extract the depth patch of the detected landmarks.
\subsection{Trilateration and Node Localization}
\begin{figure}[!ht]
\centering
\includegraphics[width=0.3\textwidth]{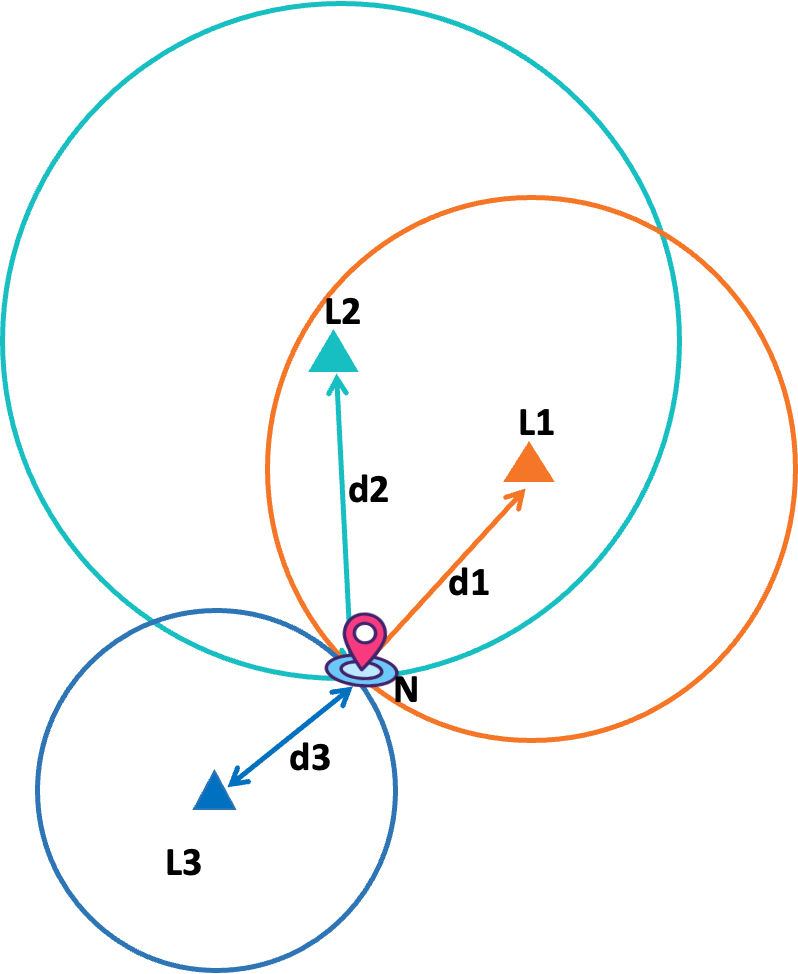}
\caption{Trilateration of Unknown Node N using Landmark Anchors} 
\label{fig_trilateration}
\end{figure}

Trilateration is a technique used to determine the position of a point by measuring the distances between that point and three known reference points as shown in Fig.\ref{fig_trilateration}.

To perform trilateration, the locations of $3$ landmark anchors are obtained using our landmark recognition model discussed in section 3.2. The distance to the landmark is calculated using the stereo vision method discussed in Section 3.3. Then, the Least square method\cite{least_square1,least_square2,least_square3} is applied to obtain the position of the unknown node (N).
Consider $L = (l1, l2... lm)$ be the set of landmark anchors with coordinates ${(x_k, y_k)}$, where $1 < k < m$ and coordinate of the unknown node is ${(x_N, y_N)}$. Then the distance from an unknown node N to each landmark in L can be represented by $D = (d1,d2...dm)$ as shown in Euclidean Eqns. \ref{distance_eqn} below.
\begin{equation}\label{distance_eqn}
    \begin{split}
        d_1^2 &= (x_1 - x_N)^2 + (y_1 - y_N)^2 \\
    d_2^2  &= (x_1 - x_N)^2 + (y_1 - y_N)^2\\
            .  .    \\             
            .  .     \\     
    d_m^2 &= (x_1 - x_N)^2 + (y_1 - y_N)^2 
    \end{split}
\end{equation}
The above m equations are processed to obtain a system of simultaneous equations of $m - 1$ dimensions, which are shown below in Eq.\ref{binary_linear_eqn}:
\begin{equation}\label{binary_linear_eqn}
    \begin{split}
        2(x_1 - x_m)x + 2(y_1 - y_m)y = {x_1}^2 + {y_1}^2 \\-({x_m}^2 + {y_m}^2) + {d_n}^2-{d_1}^2\\
        2(x_2 - x_m)x + 2(y_2 - y_m)y = {x_2}^2 + {y_2}^2\\ - ({x_m}^2 + {y_m}^2) + {d_n}^2-{d_2}^2\\
        .  . \\
        .  . \\
        2(x_{n-1} - x_m)x + 2(y_{n-1} - y_m)y = {x_{n-1}}^2 + {y_{n-1}}^2\\ - ({x_m}^2 + {y_m}^2) + {d_n}^2-{d_{n-1}}^2
    \end{split}
\end{equation}
Resolving Eq. \ref{binary_linear_eqn} in matrix from we get: AX = B. 
Then,\\
{\color{white}-}\\ 
$A =
 \begin{bmatrix}
     2(x_1 - x_m) & (y_1 - y_m) \\
	(x_2 - x_m) & (y_2 - y_m) \\
            ..\\
            ..\\
	(x_{n-1} - x_m) & (y_{n-1} - y_m) \\
 \end{bmatrix}$,\\
 {\color{white}-}\\ 
$B =
\begin{bmatrix}
    {x_1}^2 + {y_1}^2 - ({x_m}^2 + {y_m}^2) + {d_n}^2-{d_1}^2 \\
    {\color{white}-}\\ 
	{x_2}^2 + {y_2}^2 - ({x_m}^2 + {y_m}^2) + {d_n}^2-{d_2}^2 \\
    ..\\
    \\ 
    ..\\
	{x_{n-1}}^2 + {y_{n-1}}^2 - ({x_m}^2 + {y_m}^2)+ {d_n}^2\\-{d_{n-1}}^2 \\
\end{bmatrix}$,\\
$X =
\begin{bmatrix}
     x,y
\end{bmatrix}^T$ \\
{\color{white}-}\\ 
{\color{white}-}\\ 
Now the solution is obtained using the least square method in Eq.\ref{least_square_solution} where ${[.]}^T$ denotes the transpose of a matrix.
 \begin{equation}\label{least_square_solution}
    X = 
     \begin{bmatrix}
         x,y
     \end{bmatrix}^T  = (A^TA)^{-1}A^TB
 \end{equation}
 
If we select three landmark $l1,l2,l3$ from set $L$, they form circles with radii $d1, d2,$ and $d3$, respectively, and intersect at point $(x_N, y_N)$ shown in Fig.\ref{fig_trilateration}, representing the position of the unknown node N.

\subsubsection{Optimization of Estimated position:}
Trilateration can be tackled as an optimization problem considering errors in the observed distances($d_i$) to landmarks. 
By evaluating a Node's distance to each landmark anchor and comparing it to the known distances, we can assess its suitability as an approximation of the actual position. If the distances align perfectly, the point accurately represents the actual position. However, as the distances diverge, the point's proximity to the actual position decreases. Within this optimization framework, the goal is to identify the point that minimizes a specific objective function\cite{mse_optimization} and quantifies the discrepancy between the estimated distances and the known distances. Minimizing this error function leads to the most accurate approximation of the actual position. In our case, we have three sources of error, shown in Eq \ref{distance_errors} that contribute to localization error each comes from distance calculation between unknown nodes to landmarks.
\begin{equation}\label{distance_errors}
\begin{split}
     d_1 - dist\left(N, L_1\right) &= e_1\\
    d_2 - dist\left(N, L_2\right) &= e_2\\
    d_3 - dist\left(N, L_3\right) &= e_3
\end{split}
\end{equation}
A prevalent method for combining these errors is averaging their squares commonly referred to as the mean squared error (MSE) Eq.\ref{mse_equation}. This approach eliminates the possibility of negative and positive errors offsetting each other, as squares are invariably positive. So, our objective function is the mean squared error of distance between the unknown node and the landmarks considered for trilateration
\begin{equation}\label{mse_equation}
   MSE(d_i)  = \frac{\sum { \left[d_i -dist\left(N,L_i\right)\right] }^2 }{n}\\
\end{equation}
The L-BFGS-B method \cite{lbfgsOptimization}, a second-order optimization algorithm that provides an approximation of the second derivative when the direct calculation is not possible. Unlike Newton's method, which relies on the Hessian matrix, L-BFGS avoids the computational intensity of calculating the inverse of the Hessian. Instead, it approximates the inverse using the gradient, making it computationally feasible. To optimize the position of the unknown node obtained from the least square method, we minimize the above objective function(MSE) using the `L-BFGS-B' method which takes the point estimated by the least square method as the initial guess point.
\begin{multline*}
result =  minimize(MSE,initialPoint,\\
          method = `L-BFGS-B')
\end{multline*}
\section{Experiments}
\subsection{Experimental Setup}
To test the validity of our framework, a localization experiment was performed on Alienware Aurora R12 System with 11th Gen Intel Core i7 CPU, $32$ GiB Memory, and NVIDIA GeForce RTX $3070$ GPU using Python $3.10$ on PyCharm $2022.1$ \(Edu\) IDE. The landmark recognition model was trained on the Google Colab environment with $Python-3.10.12$ and $torch-2.1.0+cu118$, utilizing an NVIDIA Tesla T4 GPU with $16$ GiB memory.
\subsection{Dataset}
We used two different real-world datasets for our experiments.
\subsubsection{MSTlandmarkv1 Dataset:}
\begin{figure}[!ht]
    \centering
     \includegraphics[width=0.5\textwidth]{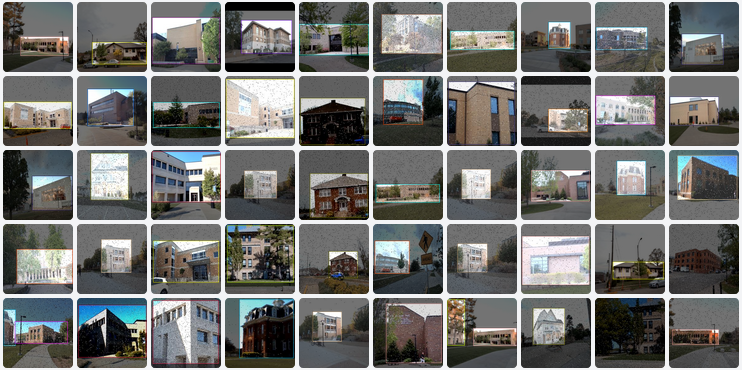}
     \caption{Example landmarks MSTlandmarkv1 dataset}
     \label{landmark_sample_data}
\end{figure}

\begin{figure}[!ht]
    \centering
     \includegraphics[width=0.48\textwidth]{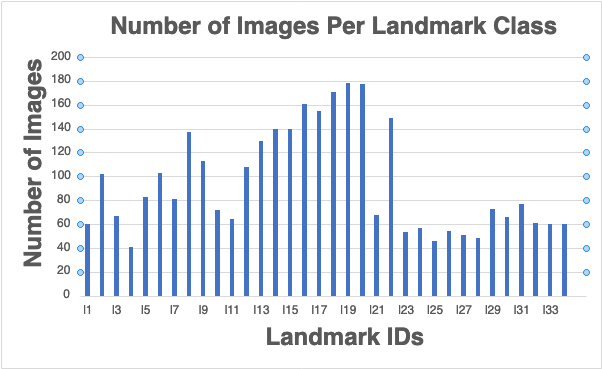}
     \caption{Landmarks image distribution classwise in MSTlandmarkv1 dataset}
     \label{image_dist_classwise}
\end{figure}

We created a real-world landmark dataset and named it the MSTlandmarkv1 dataset to train our landmark recognition model comprising around $4000$ images of $34$ different landmark instances as shown in Fig.\ref{landmark_sample_data} which were labeled manually using roboflow.ai\cite{roboflow_platform}. The image distribution per class of landmark is shown in Fig.\ref{image_dist_classwise}. The images were captured using a camera with $2K$-resolution $(2560*1440px)$. The distribution of images per landmark class is shown in Fig.\ref{landmark_stereo_examples}. After preprocessing, the dataset was split into the train($70$\% ), validation($20$\%), and test($10$\%) sets. To prevent overfitting and improve the model robustness we performed augmentation on the training images. We performed image color jittering by adjusting brightness between $-25$ percent and $+25$ percent, bounding box rotation between -15° and +15°,  and bounding box noise addition up to $5$ percent of the pixels. Finally, we obtained a dataset comprising $7547$ images maintaining the ratio of the train, validation, and test sets.
\subsubsection{MSTlandmarkStereov1 Dataset}
\begin{figure}[!ht]
    \centering
     \includegraphics[width=0.45\textwidth]{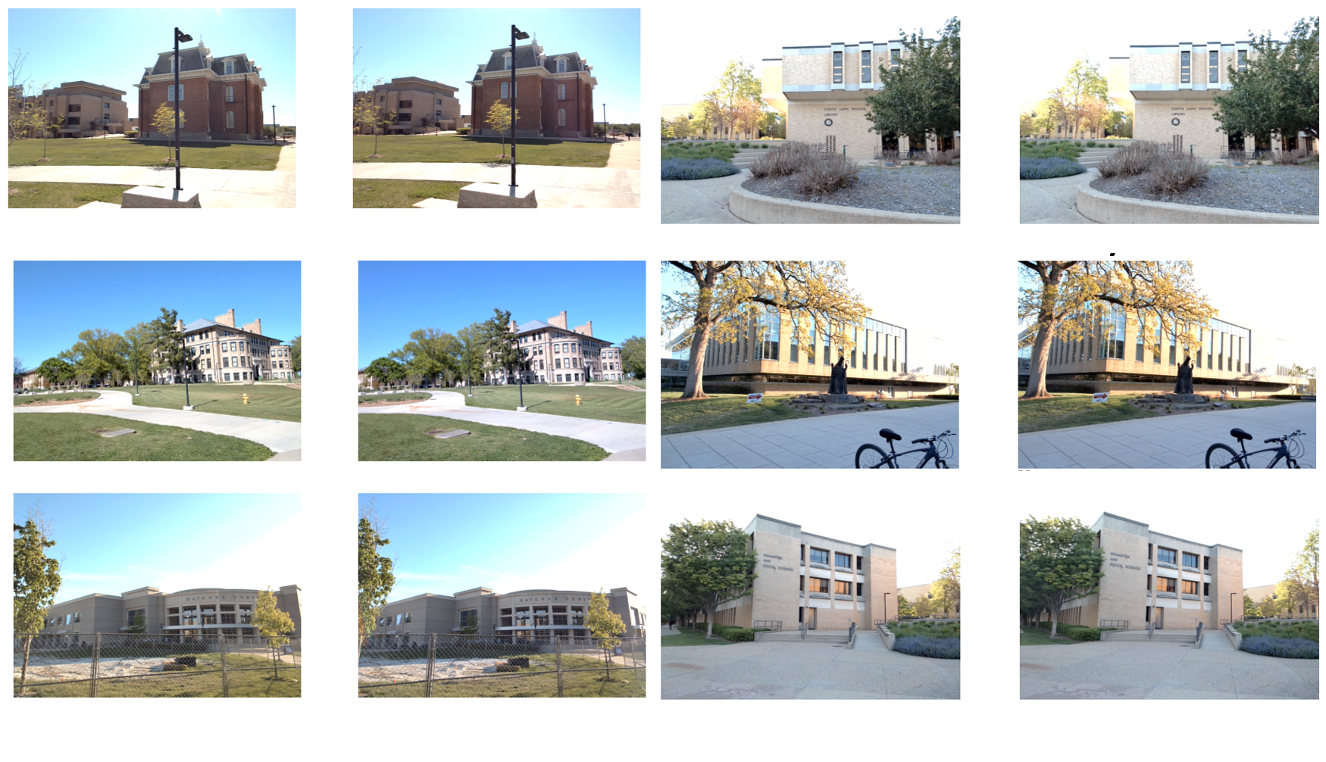}
     \caption{Example Landmark MSTlandmarkStereov1 images}
    \label{landmark_stereo_examples}
\end{figure}

\begin{figure}[!ht]
    \centering
     \includegraphics[width=0.4\textwidth]{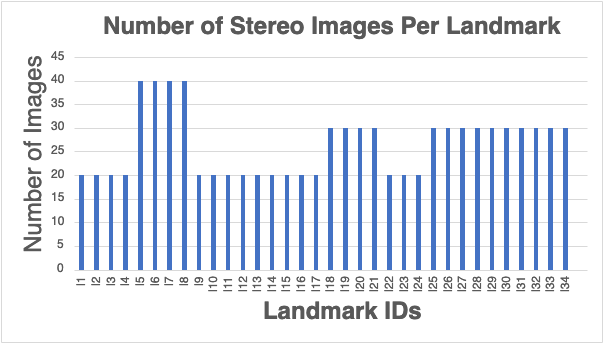}
     \caption{Landmarks stereo image distribution in MSTlandmarkStereov1 dataset}
     \label{stereo_image_dist}
\end{figure}

\begin{figure}
\includegraphics[width=0.45\textwidth]{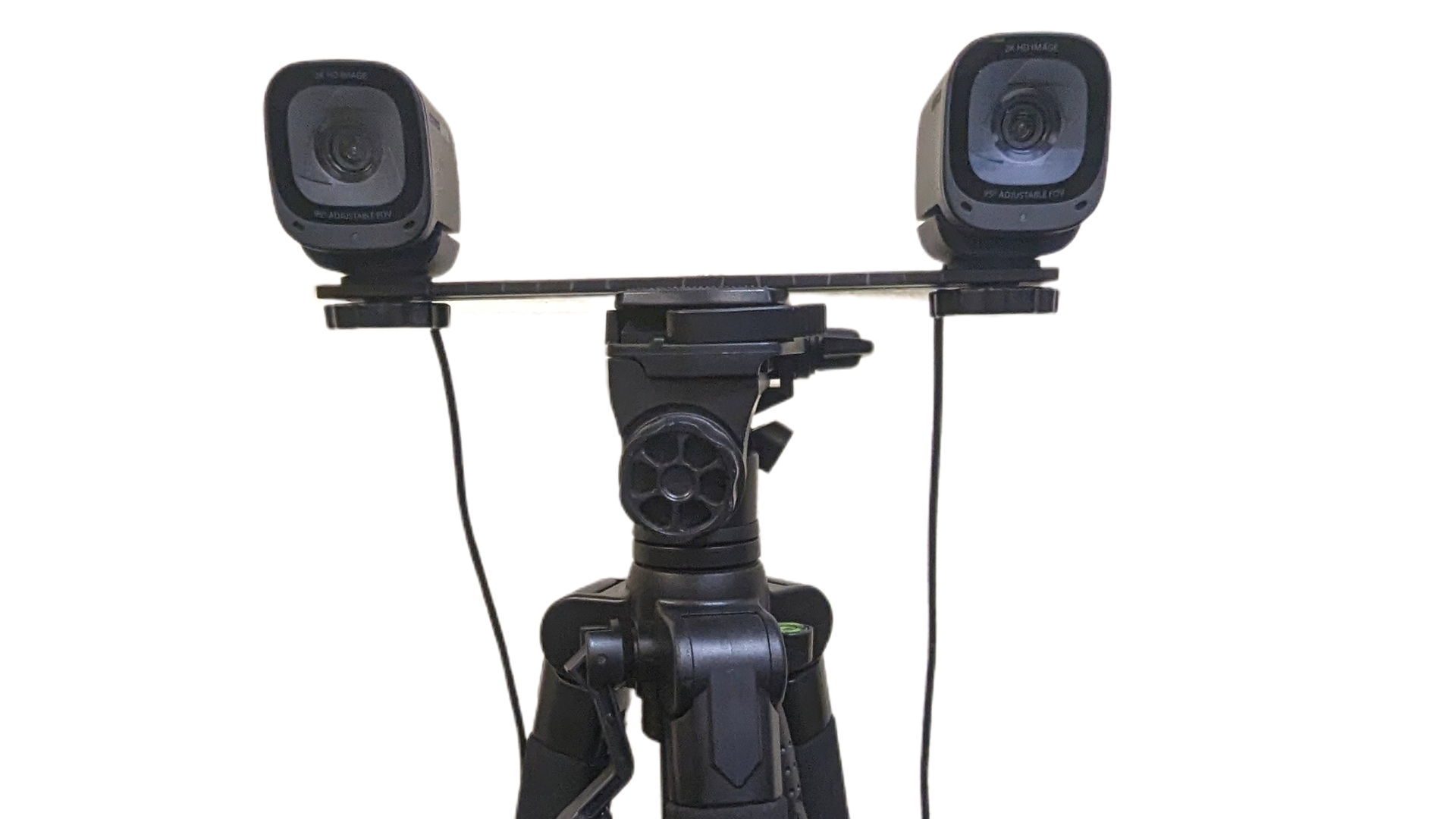}
\centering
\caption{Custom Stereo Camera Setup for Node to Landmark Distance Measurement} 
\label{custom_stereo_rig}
\end{figure}

We created another real-world landmark stereo dataset shown in Fig.\ref{landmark_stereo_examples} using our stereo image capture framework. We used two identical cameras of $(2560*1440px~@~30~fps)$ resolution as shown in Fig. \ref{custom_stereo_rig} with adjustable field of view (FoV). We arranged the camera on the long metal bar that allows us to adjust the baseline $(B)$ from $10-40$ centimeters. The stereo image distribution per landmark class is shown in Fig.\ref{stereo_image_dist}. This dataset is used to perform distance estimation experiments from unknown nodes to the landmark anchors. Using this dataset, we tested the distance estimation performance of our framework with other state-of-the-art methods.

\subsection{Trainnig Model}
We took the model $YOLOv8.0.20$ from Ultralytics\cite{yolov8_github} and trained it using our MSTlandmarkv1 dataset. We adjusted the number of epochs and modified the stack size to train the upper layers of the model and to learn and classify $34$ different classes of landmarks present in our dataset. Training a model for $30$ epochs took $0.720$ hours. The hyperparameters used to train the model are shown in Table \ref{training_hperparameters}.
\begin{table}[!ht]
\centering
\begin{tabular}{|l|l|}
\hline
\textbf{Hyperparameters} & \textbf{Values} \\ \hline
Optimizer                & SGD             \\ \hline
Learning Rate(lr) & 0.01                    \\ \hline
Warmup Momentum          & 0.8             \\ \hline
Momentum(Beta1)          & 0.937           \\ \hline
Weight Decay             & 0.001           \\ \hline
Warmup Epochs            & 3.0               \\ \hline
Epoch                    & 30              \\ \hline
Batch Size               & 16              \\ \hline
\end{tabular}
\vspace{0.1in}
\caption{Hyperparameters used to Train the model}
\label{training_hperparameters}
\end{table}
\subsubsection{Training Results and Discussion}
\begin{figure}[!ht]
\centering
\includegraphics[width=0.5\textwidth]{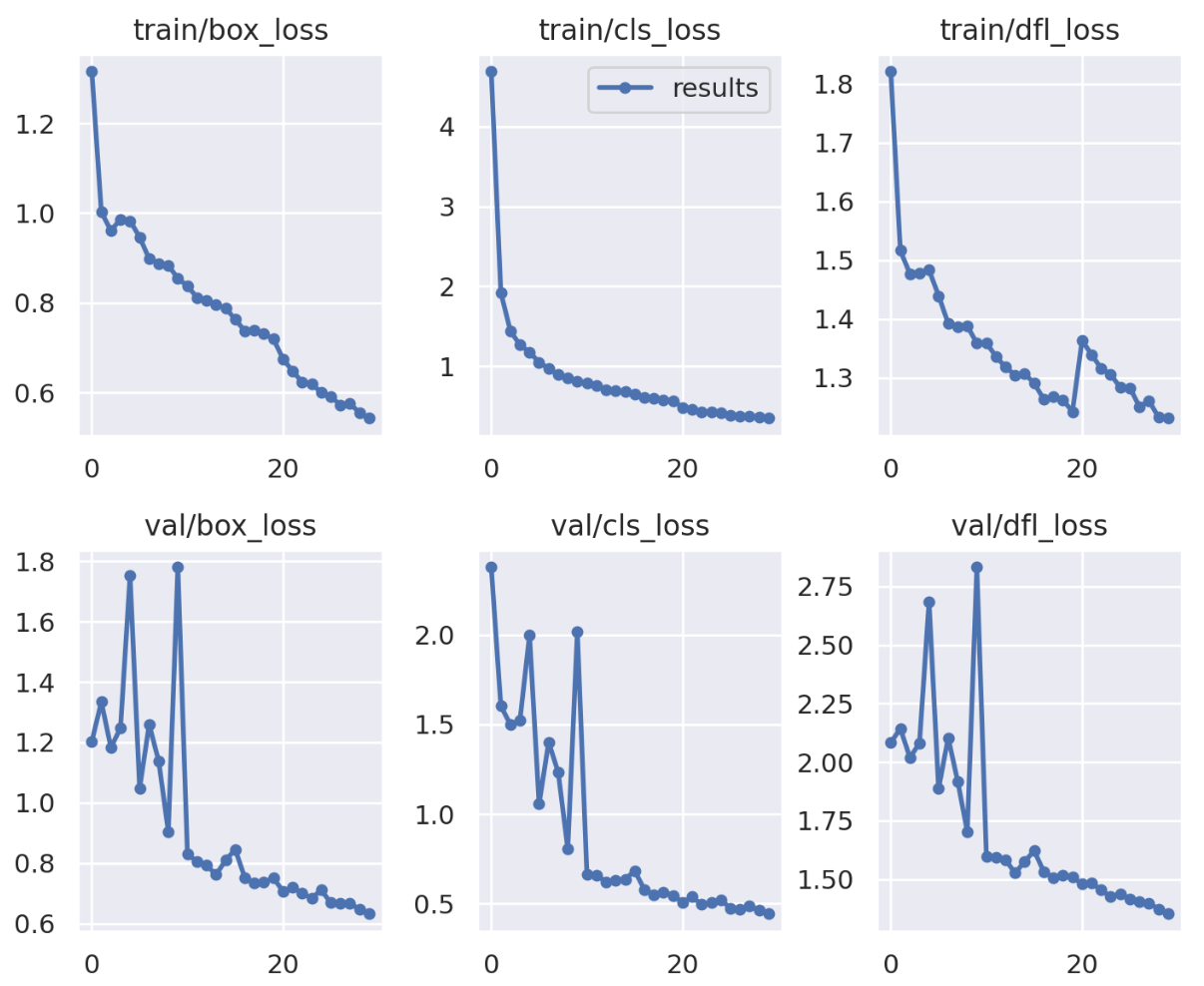}
\caption{Graph showing box loss, objectness loss, classification loss over the training epochs for the training and validation} 
\label{fig_train_result1}
\end{figure}

\begin{figure}[!ht]
\centering
\includegraphics[width=0.5\textwidth]{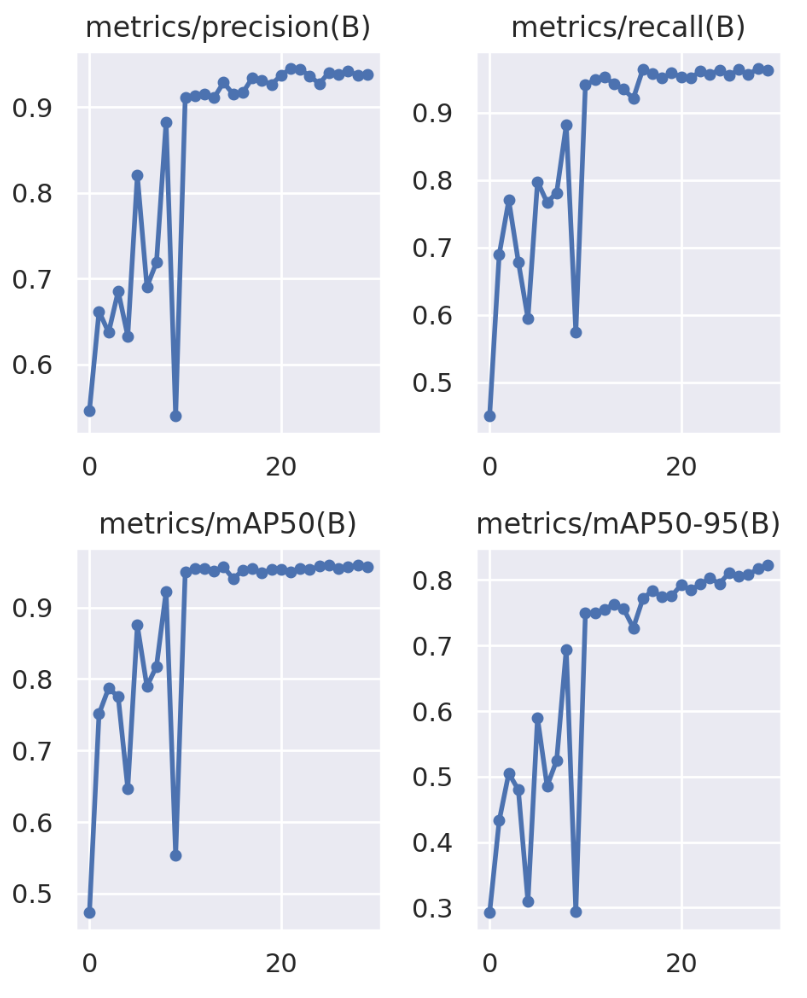}
\caption{Graph showing precision, recall, and mean average precision $(mAP)$ over the training epochs for the training and validation} 
\label{fig_train_result2}
\end{figure}

Fig. \ref{fig_train_result1} represents graphs showing how box loss, objectness loss, and classification loss are improving over the training Epochs. Fig.\ref{fig_train_result2} represents the model performance scores (Precision, Recall, and mAP) as a result of training and validation. Box loss represents how well the predicted bounding box aligns with the ground truth bounding box. Objectness Loss typically represents the likelihood that an object is present in a proposed region. This helps the model distinguish between regions containing objects and those that do not. Classification loss measures how well the model predicts the class of the detected object.
The model's performance improved significantly in terms of precision, recall, and mean average precision (mAP) until it reached a stable point after about 20 epochs. The box, objectness, and classification losses for the validation data also decreased rapidly until around epoch $20$. Therefore, we employed early stopping to select the optimal weights. Our model shows a Box loss precision(Pr) of $0.939$, a Recall (R) of $0.963$, and achieved a mean Average Precision $(mAP~@~0.5~IoU)$ score of $0.957$ and $mAP~@~[0.5:0.95] ~ IoU$ of $0.823$ for all classes.
\subsection{Distance Measurement Using Stereo Vision}
For distance measurement from the unknown node position to the landmark anchors, we applied the SGM Algorithm \ref{alg_SSD} for disparity map generation and applied triangulation using Eq. to obtain the depth map. The depth patch containing the landmark anchor is cropped using bounding box coordinates obtained from landmark recognition result which is aggregated to obtain an estimated distance from a node carrying the stereo camera. The actual distance from a node to the landmark was calculated using the great-circle distance method which computes the distance between two geographic coordinates(landmark anchor position and node position in our case) on the surface of the earth. The actual node position was captured and stored while creating the MSTlandmarksterov1 dataset. Landmark anchor location is obtained from a pre-defined landmark database.

\subsection{Localization Experiment and Results}
We obtained different sets of results from our localization experiment based on MSRlandmarkv1 and MSTlandmarkstereov1. We performed our localization experiment on eight different trilateration sets, each consisting of 3 landmark anchors which were located within the range of 38-78 meters with unknown nodes. we also obtained results for SLAM and VO methods that use the KITTI benchmark stereo dataset\cite{kittiSLAM,kittiVO}, and compared them with our results.
\begin{figure}
\includegraphics[width=0.4\textwidth]{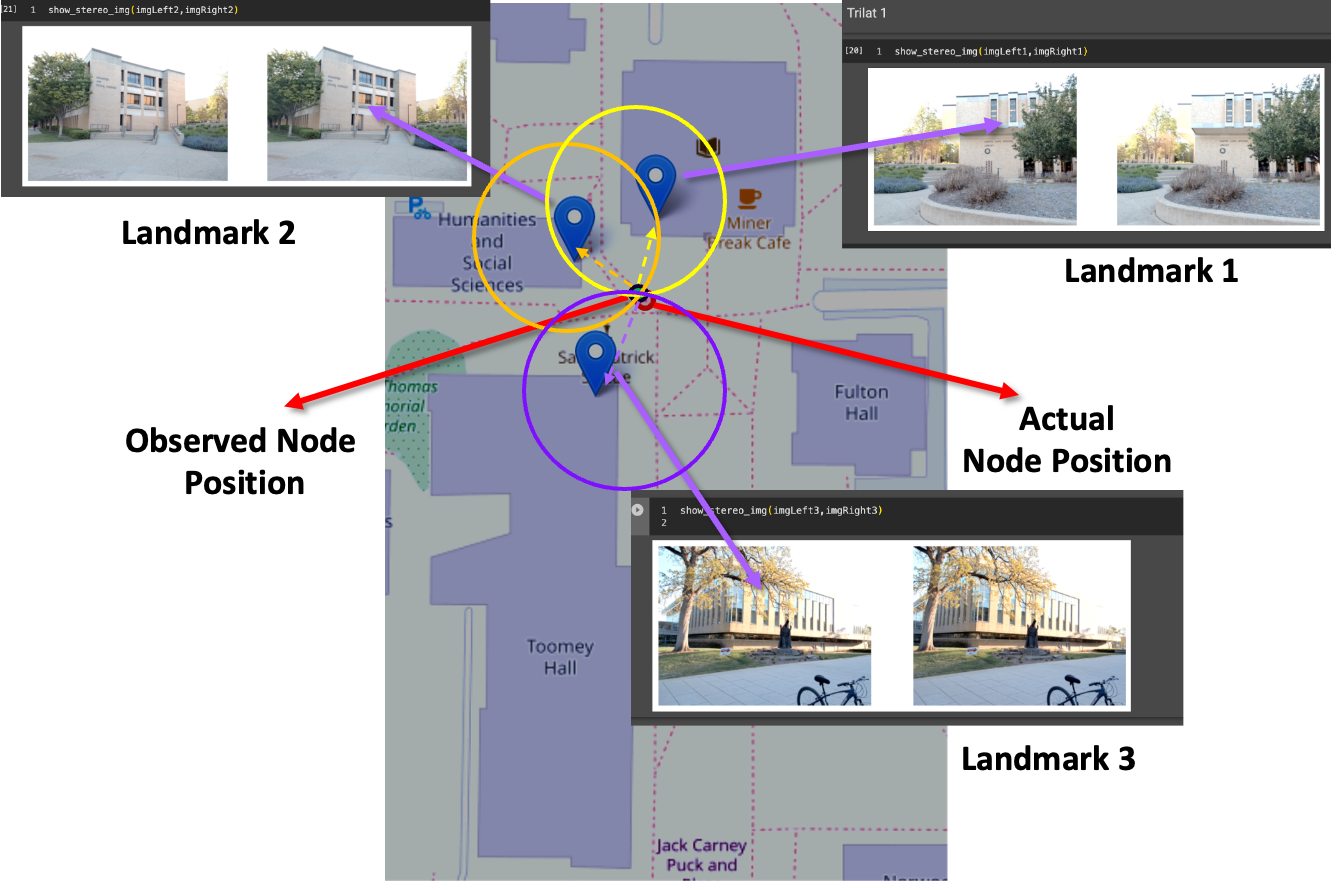}
\centering
\caption{Distance Measurement Example} 
\label{localization_example1}
\end{figure}
\subsubsection{Comparison with Visual Odometry and SLAM Based Methods}
\begin{figure}[!ht]
    \includegraphics[width=0.5\textwidth]{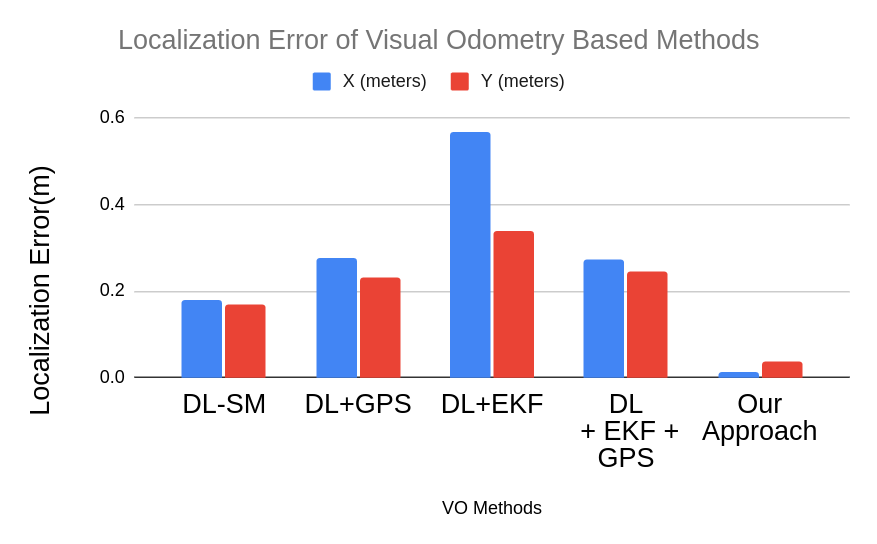}
    \caption{VO Methods}
    \label{vo_graph}
\end{figure}

\begin{figure}[!ht]
        \includegraphics[width=0.5\textwidth]{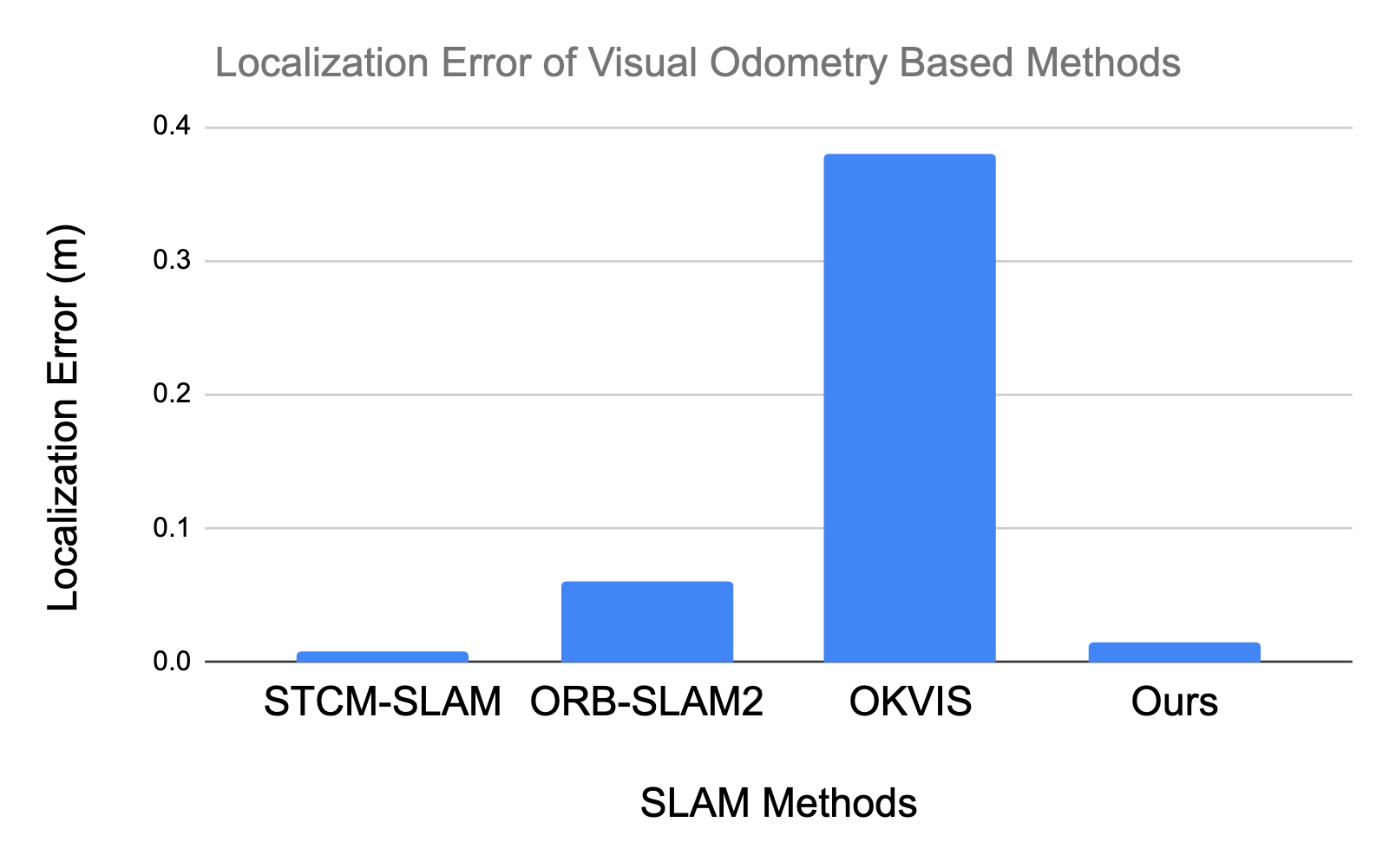}
        \caption{SLAM Methods}
        \label{slam_graph}
\end{figure}

We compared different SOTA visual odometry(VO) and Simultaneous Localization and Mapping (SLAM) based approaches in terms of translation error(along X coordinates and y coordinates) tested on \cite{kittiSLAM,kittiVO}. Our approach (tested on MSTlandmarkstereov1) shows outstanding performance in terms of localization error (RMSE) compared to other VO and SLAM-based methods. Inference results of VO-based Deep localization approaches by \cite{deepLocalization} show different localization errors when they are assisted with synthetic measurements, GPS, and Kalman Filter separately and together. DeepLocalization assisted with synthetic measurement shows minimal error along the x and y coordinates. However, our approach shows minimal localization error without the assistance of any GPS or filter. Results are shown in Table \ref{voComaprision} and Fig.\ref{vo_graph}. Results of the Localization of SLAM-based methods are shown in Table \ref{slamComaprision} and Fig.\ref{slam_graph}. Our average localization error(RMSE) is better compared to the overall pose estimation error of ORB-SLAM2 \cite{reforbslam2} and the OKVIS \cite{OKVISSLAM2} method. However, it is more comparable and closer to SCTM-SLAM which incorporates semantic segmentation and contextual information for pose estimation.
\begin{table}[!ht]
\centering
\begin{tabular}{|l|l|l|}
\hline
Visual Odometry Methods & RMSE X (m)  & RMSE Y (m)  \\ \hline
\begin{tabular}[c]{@{}l@{}}DeepLocalization:\\synthetic measurements \end{tabular}  
  &{0.178} & {0.17} \\ \hline
\begin{tabular}[c]{@{}l@{}}DeepLocalization\\+ GPS-based Inference   \end{tabular} & 0.276 & 0.231 \\ \hline
\begin{tabular}[c]{@{}l@{}}DeepLocalization +\\  Filter-based Inference\end{tabular} & 0.566 & 0.339\\ \hline
\begin{tabular}[c]{@{}l@{}}DeepLocalization\\  + EKF + GPS\end{tabular}  & 0.271 & 0.245 \\ \hline
\textbf{Our Approach}                                                                & \textbf{0.0142}           & \textbf{0.039}           \\ \hline
\end{tabular}
\vspace{0.1in}
\caption{Comparing Localization errors with Different Visual Odometry(VO) methods based on KITTI dataset}
\label{voComaprision}
\end{table}

\begin{table}[!ht]
\centering
\begin{tabular}{|l|l|}
\hline
Methods      & RMSE Overall(m)           \\ \hline
STCM-SLAM    & 0.008                     \\ \hline
ORB-SLAM2    & {0.06}                    \\ \hline
OKVIS        & 0.379                     \\ \hline
Our Approach & \textbf{0.0147}          \\ \hline
\end{tabular}
\vspace{0.1in}
\caption{Comparing Localization errors with Different SLAM based Localization approach based on KITTI Dataset}
\label{slamComaprision}
\end{table}

\subsubsection{Comparison with DV Hop Based method}
\begin{figure}[!ht]
\includegraphics [width=0.5\textwidth]{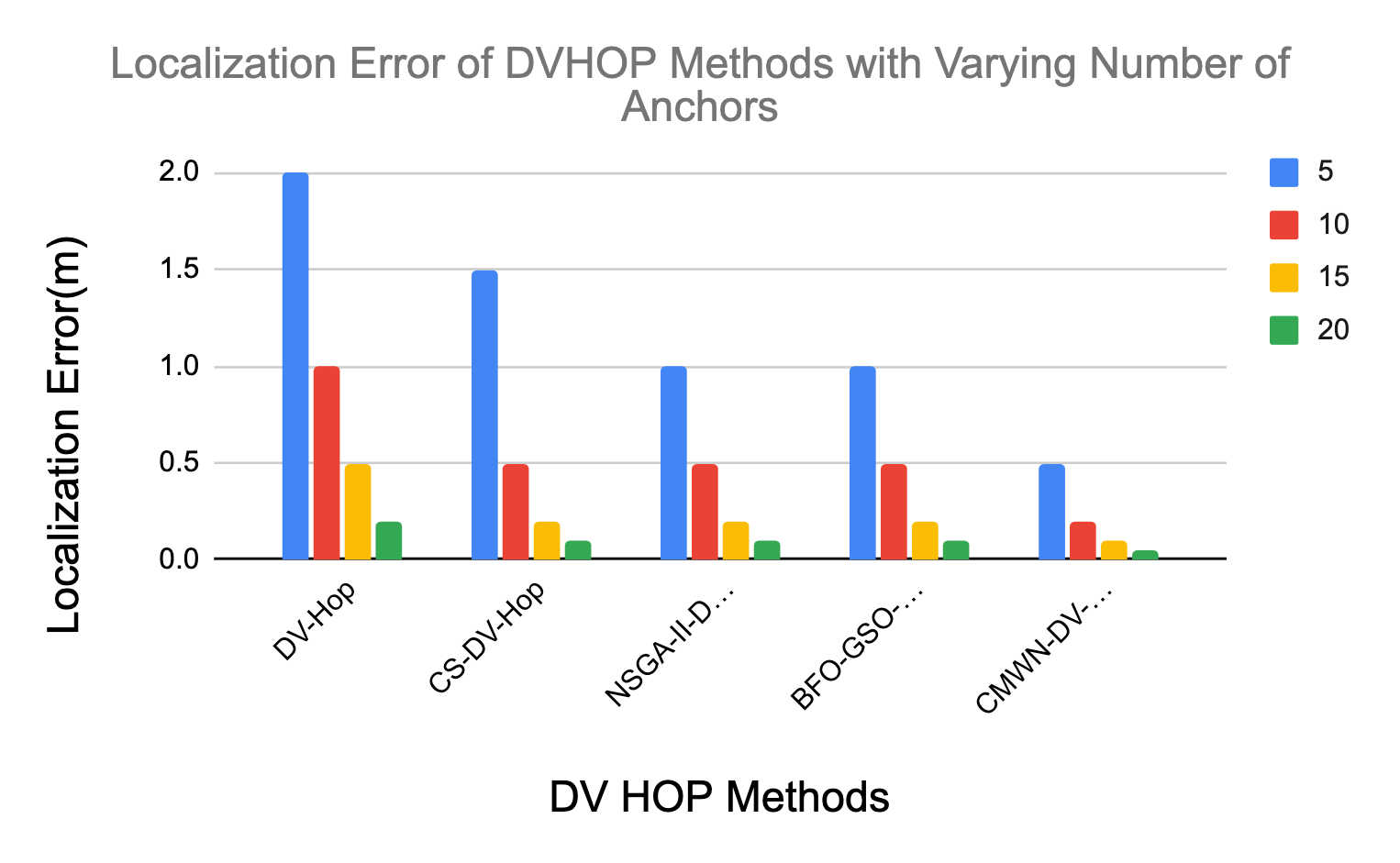}
\centering
\caption{Localization error of Different DV HOP algorithms with varying Anchors} 
\label{dv_hop_graph}
\end{figure}

Our approach uses three landmark anchors for trilateration and gives localization RMSE of $\textbf{0.0146921 m}$. Localization error range (in meters) of DV hop-based Localization algorithms\cite{dvhop_aps,csdvhop,dvhop_nsgaii,dv_hop_Cyclotomic} are shown in Table \ref{dvhopComparision}. Fig.\ref{dv_hop_graph} shows that the DV-Hop method gives a higher localization error when the Number of Anchors is $5$ and gradually improves as the number of anchors increases.  The CMWN-DV Hop shows the smallest error range (0.05-0.2) meters compared to other DV-Hop methods(NSGA-II DV, CS-DV, and Standard DV-Hop) when the number of anchors is increased to 20. However, our average error is still minimal compared to CMWN-DV.
\begin{table}[!ht]
\centering
\begin{tabular}{|l|l|l|l|l|}
\hline
Anchors & DV-Hop & CS-DV & NSGA-II-DV & CMWN-DV \\ \hline
5              & 2-3    & 1.5-2     & 1-2           & 0.5-1 \\ \hline
10             & 1-2    & 0.5-1.5   & 0.5-1         & 0.2-0.5 \\ \hline
15            & 0.5-1   & 0.2-0.8   & 0.2-0.5       & 0.1-0.3 \\ \hline
20            & 0.2-0.5 & 0.1-0.4   & 0.1-0.3       & 0.05-0.2 \\ \hline
\end{tabular}
\vspace{0.1in}
\caption{Comparing localization errors range(meters) of different DV-HOP based localization approach with varying number of anchor nodes}
\label{dvhopComparision}
\end{table}

\section{Conclusion and Future Works}
The proposed framework provides a novel and effective approach for localization in non-GPS battlefield environments using passive camera sensors and naturally existing or artificial landmarks as anchors. The method utilizes a custom-calibrated stereo-vision camera and the YOLOv8s Model to obtain the location of landmark anchors. The disparity maps of detected landmark stereo pairs are generated and distances to landmarks are determined by extracting the depth patch utilizing a bounding box obtained from the recognition model. The least square method is applied to obtain the coordinates of unknown nodes, and the L-BFGS-B optimization algorithm is employed to achieve a more accurate node position. Experimental results demonstrate that the proposed method outperforms existing anchor-based DV-Hop algorithms and vision-based algorithms in terms of localization error(RMSE). We aim to expand this methodology to address the localization needs of maneuvering forces in battlefield areas where GPS signals are unavailable or deliberately denied. This extension will involve the development of algorithms for safe path planning and the avoidance of potential hazards, considering the positions of landmarks and the locations of enemy forces.

\end{document}